\newcommand{\CLASSINPUTtoptextmargin}{0.75in}
\newcommand{\CLASSINPUTbottomtextmargin}{1.1in}

\documentclass[10pt,conference]{IEEEtran}
\IEEEoverridecommandlockouts

\usepackage{cite}
\usepackage{amsmath,amssymb,amsfonts}
\usepackage{algorithmic}
\usepackage{graphicx}
\usepackage{textcomp}
\usepackage{xcolor}
\usepackage[ruled,vlined]{algorithm2e}
\usepackage{tabularx,booktabs}
\usepackage{threeparttable}
\usepackage{url}
\usepackage{placeins}  
\usepackage{stfloats}  

\def\BibTeX{{\rm B\kern-.05em{\sc i\kern-.025em b}\kern-.08em
    T\kern-.1667em\lower.7ex\hbox{E}\kern-.125emX}}
\begin{document}
\columnsep 0.24in
\title{LLM-FS: Zero-Shot Feature Selection for Effective and Interpretable Malware Detection
}



\author{
\IEEEauthorblockN{Naveen Gill\textsuperscript{*,a}, Ajvad Haneef K\textsuperscript{*,b}, and S. D. Madhu Kumar\textsuperscript{*,c}}
\IEEEauthorblockA{
\textit{\textsuperscript{*}Department of Computer Science and Engineering} \\
\textit{\textsuperscript{*}National Institute of Technology Calicut, India} \\
Emails: \{\textsuperscript{a}gillnaveen121@gmail.com, \textsuperscript{b}ajvad\_p210054cs@nitc.ac.in, \textsuperscript{c}madhu@nitc.ac.in\}
}
}

\maketitle
\begin{abstract}
Feature selection (FS) remains essential for building accurate and interpretable detection models, particularly in high-dimensional malware datasets. Conventional FS methods such as Extra Trees, Variance Threshold, Tree-based models, Chi-Squared tests, ANOVA, Random Selection and Sequential Attention rely primarily on statistical heuristics or model-driven importance scores, often overlooking the semantic context of features. Motivated by recent progress in LLM-driven FS, we investigate whether large language models (LLMs) can guide feature selection in a zero-shot setting, using only feature names and task descriptions, as a viable alternative to traditional approaches. We evaluate multiple LLMs (GPT-5.0, GPT-4.0, Gemini-2.5 etc.) on the EMBOD dataset (a fusion of EMBER and BODMAS benchmark datasets), comparing them against established FS methods across several classifiers, including Random Forest, Extra Trees, MLP, and KNN. Performance is assessed using accuracy, precision, recall, F1, AUC, MCC, and runtime. Our results demonstrate that LLM-guided zero-shot feature selection achieves competitive performance with traditional FS methods, while offering additional advantages in interpretability, stability, and reduced dependence on labeled data. These findings position zero-shot LLM-based FS as a promising alternative strategy for effective and interpretable malware detection, paving the way for knowledge-guided feature selection in security-critical applications.
\end{abstract}
\begin{IEEEkeywords}
Machine Learning, Feature Selection, Large Language Models (LLMs), Malware Detection, Cyber Security
\end{IEEEkeywords}

\section{Introduction}
The rapid advancement of Large Language Models (LLMs) has transformed diverse domains, including natural language processing, healthcare, and more recently, cybersecurity\cite{raiaan2024review}. By leveraging vast pretraining corpora and effective prompting strategies, LLMs demonstrate impressive few-shot and zero-shot capabilities, enabling them to generalize across tasks without requiring explicit task-specific training. These capabilities have motivated researchers to explore LLMs beyond traditional language tasks, applying them to structured tabular problems such as feature selection and engineering.
Feature selection (FS) plays a central role in building efficient, accurate, and interpretable models, especially in high-dimensional domains such as malware detection\cite{hasan2025enhancing,rasheed2023impact, fang2019feature}. Conventional FS methods—including filter-based approaches (e.g., Chi-Squared tests, ANOVA, Variance Threshold), wrapper-based approaches (e.g., Sequential Feature Selection), and embedded approaches (e.g., Tree-based models, LASSO)—have proven effective in reducing dimensionality and improving model performance\cite{barbieri2024analysis,liyew2025review}. However, these approaches suffer from key limitations including dependence on statistical heuristics or model-driven scores, instability across runs, and strong reliance on labeled data\cite{khaire2022stability}. In contrast, LLM-guided feature selection offers a knowledge-driven alternative. By exploiting the semantic understanding encoded within pretrained LLMs, feature importance can be inferred through zero-shot prompting without requiring access to raw data distributions Recent works such as LLM-Select\cite{jeong2024llm} demonstrate that even when provided with only feature names and task descriptions, LLMs can identify predictive features with performance competitive to data-driven methods like LASSO Similarly, frameworks such as FREEFORM\cite{lee2025knowledge} and LLM-FE\cite{abhyankar2025llm} highlight that LLMs bring additional benefits in terms of interpretability, stability, and knowledge integration across domains like genetics and tabular learning.
These insights motivate the exploration of LLM-based FS in malware detection, where both dimensionality reduction and interpretability are critical for practical deployment.\\
Malware detection datasets, such as EMBER\cite{anderson2018ember} and BODMAS\cite{yang2021bodmas}, present a particularly challenging testbed: they contain hundreds of features, are highly imbalanced, and require careful feature selection to avoid overfitting while maintaining generalizability. While conventional FS methods have been widely applied to these datasets, the potential of LLM-guided FS remains underexplored. Furthermore, existing studies on LLM-based FS have primarily focused on biomedical or anomaly-detection tasks, leaving a gap in large-scale, high-dimensional cybersecurity applications\cite{santana2024stacking, wang2024initial}. In this paper, we propose LLM-FS framework to address this gap by investigating LLM-guided zero-shot feature selection for malware detection. We systematically compare LLM-based strategies (using GPT-4.0, GPT-4.0-mini, and Gemini-2.5) against a wide range of established FS methods, including Extra Trees, Variance Threshold, Tree-based importance, Chi-Squared, ANOVA, Random Selection, and Sequential Attention\cite{barbieri2024analysis, pudjihartono2022review}. Experiments are conducted on the EMBOD dataset, a fusion of EMBER and BODMAS benchmarks, with multiple classifiers including Random Forest, Extra Trees, MLP, and KNN\cite{akhtar2022malware}.\\
Our results demonstrate that LLM-FS achieves performance comparable to traditional methods while offering additional advantages in interpretability, stability across runs, and reduced dependence on labeled training data. This establishes LLM-FS as a viable and complementary alternative to conventional techniques, particularly for large-scale, high-dimensional malware datasets.\\
The key contributions of this work are:
\begin{itemize}
    \item We present LLM-FS, a zero-shot feature selection framework leveraging LLMs for effective and interpretable malware detection.
    \item Provide a comprehensive evaluation of LLM-FS methods against classical FS techniques across multiple classifiers on the EMBOD dataset.
    \item Establish the LLM-FS as an alternative paradigm, reducing reliance on labeled data and providing a foundation for future research in large-scale malware detection.
\end{itemize}
The rest of the paper is organized as follows: Section II presents an overview of related works in the field of LLM-based FS. Section III details the methodology employed in the LLM-FS framework, which includes feature selection and classifier model training. Section IV presents the experimental results, the evaluation metrics used to assess the performance of the LLM-FS, and a discussion of future directions for research in this area. Finally, section V summarizes the paper and provides the conclusion.

\section{Related Works}

\subsection{Feature Selection in Machine Learning}
Feature selection (FS) is a fundamental step in building efficient and interpretable machine learning models. Traditional FS methods can be broadly categorized into filter, wrapper, and embedded approaches. Filter methods (e.g., Variance Threshold, Chi-Squared, ANOVA) rely on statistical measures or information-theoretic scores to rank features independently of the learning algorithm. Wrapper methods (e.g., Sequential Feature Selection) iteratively evaluate subsets of features using a classifier, but are computationally expensive. Embedded methods (e.g., LASSO, Tree-based models, Extra Trees) integrate feature selection directly into the training process, typically through sparsity-inducing regularization or feature importance measures. While effective, these methods often lack interpretability, may be unstable across runs, and depend heavily on labeled data.

\subsection{LLMs for Feature Selection}
The emergence of large language models (LLMs) has motivated a new line of research in feature selection that leverages semantic reasoning and prior knowledge. LLM-Select\cite{jeong2024llm} demonstrated that LLMs, when prompted with only feature names and task descriptions, can identify predictive features with performance comparable to traditional methods such as LASSO . Their study introduced multiple prompting strategies (LLM-Score, LLM-Rank, and LLM-Seq) and showed that zero-shot LLM-based FS can rival data-driven baselines in healthcare, finance, and social science tasks. Similarly, Exploring LLMs for Feature Selection categorized LLM-FS methods into data-driven (requiring numerical values) and text-based (leveraging semantic associations), emphasizing the promise of text-based approaches for real-world applications.
Advancing this idea, Knowledge-Driven Feature Selection (FREEFORM)\cite{lee2025knowledge} proposed a framework combining chain-of-thought reasoning and ensembling for genetic data, showing better performance in low-data systems. Moreover, LLM-FE\cite{abhyankar2025llm} dived into automated feature engineering with evolutionary optimization led by LLMs, displaying that knowledge-driven reasoning can advance programmatic feature discovery for tabular datasets. Also, ensemble-based techniques have been proposed to minimize the variability of single LLM outputs\cite{santana2024stacking}. The authors investigated the use of multiple LLMs for anomaly detection, where each model contributed a feature subset and their results were merged using stacking methods. These works highlight the growing acknowledgment of LLMs as a valuable tool for FS, either as solo selectors or as part of hybrid pipelines.

Even after these progresses, many limitations persist. First, most existing studies stress on biomedical or anomaly detection tasks, with less focus on cybersecurity and malware detection, where high-dimensional datasets and rapidly changing feature distributions pose unique threats. Second, while prior works show more importance to accuracy, relatively few research studies systematically judge interpretability, stability, and runtime efficiency of LLM based FS techniques, factors that are particularly important in security-critical systems. Lastly, even though LLM-Select has displayed that zero-shot prompting is quite effective, its usability to large-scale malware datasets such as EMBER, BODMAS, and EMBOD stays untouched. In pointing these voids, our work gives the first comprehensive evaluation of LLM-guided zero-shot FS in the subject of malware detection. We compare multiple LLMs (GPT-4.0, GPT-4.0-mini, Gemini-2.5) against a plethora of traditional FS methods across multiple classifiers on the EMBOD dataset. Our findings show that LLM-FS gets competitive performance with traditional methods, while giving special advantages that make it a trustable alternative framework for security-critical, high-dimensional applications.
\section{Methodology}
\begin{figure*}[h!]
\centering
  \includegraphics[width=16cm]{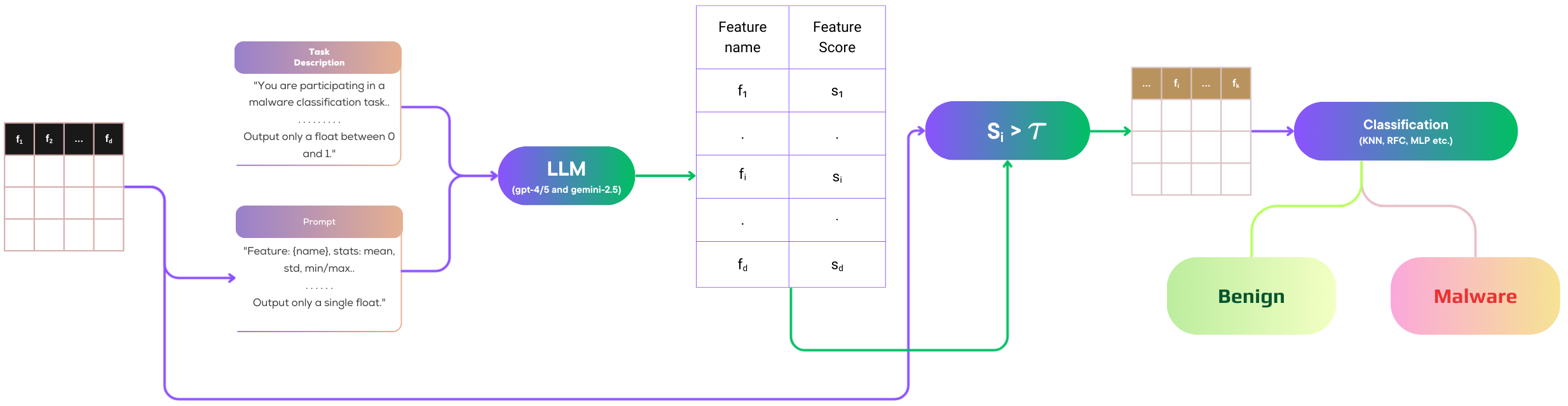}
  \caption{Architecture of proposed LLM-FS framework}
  \label{fig:architecture}
\end{figure*}

The block diagram of the proposed LLM-FS framework is shown in Figure \ref{fig:architecture}. LLM-FS makes use of a prompt-based approach, depending on a feature’s name and statistical properties, LLM gives independent scores. The scores are then combined to pick the top-k features for training different classifiers. The main innovation is in the feature selection step, where we compare traditional FS methods with LLM-led zero-shot techniques. The chosen features are then used to train various classifiers, and performance is judged using a comprehensive set of metrics. The following subsections describe the main components of the paradigm.

\subsection{Feature Engineering and Selection}
Feature engineering involves converting raw static and dynamic behavioral features into a well-ordered depiction suitable for training machine learning models. Many times these features are represented in high-dimensional spaces, harnessing various details about malware behavior, such as API call frequencies, file metadata, opcode sequences, and network activity patterns. FS has an important role in minimizing the drawbacks of dimensionality, reducing overfitting, and boosting model efficiency and explainability.\\
Let $n$ be the number of samples and $d$ be the number of features in the dataset, and $y$ be the binary class labels showing if a sample is benign (0) or malware (1).
The dataset can be represented as:
\begin{equation}
  X \in \mathbb{R}^{n \times d}, \quad y \in \{0,1\}^n
\end{equation}  
In malware detection, the aim of feature selection is to identify a reduced feature matrix
\begin{equation}
X' \in \mathbb{R}^{n \times k}, \quad k < d,
\end{equation}  
It keeps classification performance intact or improves it while reducing extra parts and computational overhead. Formally, the feature selection task can be written as:
\begin{equation}
X' = \underset{X' \in \mathbb{R}^{n \times k}, \; k < d}{\arg\max} \; \mathcal{M}(X', y),
\end{equation}  

where $\mathcal{M}$ represents a performance metric such as Accuracy, F1-score, AUC, or MCC of a classifier trained on $X'$, as indicated in section \ref{sec:metrics}.
\subsubsection{FS using Traditional Methods}
Traditional FS methods can be broadly categorized into filter, wrapper, and embedded approaches. These methods relies on statistical heuristics or model-driven importance scores to rank and select features. The following traditional FS methods employed in this study:
\begin{itemize}
    \item \emph{Variance Threshold:}  
    This method eliminates features whose variance falls below a user-defined threshold, under the assumption that low-variance features contribute little to distinguishing between malware and benign samples\cite{guyon2003introduction}. For example, a feature that takes nearly constant values across the dataset provides almost no information for classification. Although simple and computationally efficient, this method may discard features that are rare but highly discriminative.  

    \item \emph{Chi-Squared Test ($\chi^2$):}  
    The chi-squared test evaluates the statistical dependence between individual features and the class label \cite{liu1995chi}. 
    A bigger value of $\chi^2$ shows a prominent association, giving the idea that the feature is more relevant for bifurcating between malware and benevolent classes. This method is very particular with categorical or discretized features and also very effective, but it asserts non-negativity and may fail to capture nonlinear dependencies.  

    \item \emph{ANOVA F-Test:}  
    The Analysis of Variance (ANOVA) F-test measures the importance of differences in feature means across classes by calculating the ratio between-class variance to within-class variance \cite{lakshmaiah2021anovaids}. Features with greater F-scores are known to be more discriminative. While effective for continuous features, ANOVA already assumes normality and homoscedasticity, which may not be true for all malware datasets.

    \item \emph{Mutual Information (MI):} 
  MI quantifies the amount of information relayed between a feature and the class label, accounting for both linear and nonlinear dependencies\cite{peng2005feature}. A higher MI value depicts a stronger relationship, showing that the feature is more informative for classification. MI is multifunctioning and can handle various feature types, but guessing MI accurately can be challenging, especially in high-dimensional spaces with bounded data.

    \item \emph{Correlation Thresholding:}
    The correlation thresholding method identifies and removes repeated features based on their pairwise correlation coefficients, usually using Pearson’s correlation \cite{hall1999correlation}. Features that shows high correlation (above a defined threshold) with other features are considered redundant and are removed from the feature set. Highly inter-correlated features often provide repeated information, which can lead to multicollinearity and decrease model interpretability. 

    \item \emph{Tree-Based Importance:}  
    Tree-based FS builds decision tree ensembles to judge feature importance depending on their contribution to reducing malcontent in the classification task\cite{breiman2001random}. Features that regularly reduce malcontent across trees are known to be more important. This technique is efficient in managing feature interactions and can fit nonlinear decision boundaries, making it suitable for sophisticated malware datasets.

    \item \emph{ExtraTrees Classifier:}  
    Extremely Randomized Trees (ExtraTrees) is a tree-based ensemble approach that can be used for feature selection\cite{geurts2006extremely}. It brings more randomness in comparison to traditional RF by choosing random split points for every feature, instead of searching for the optimal split. ExtraTrees are particularly very effective for high-dimensional datasets, where randomization facilitates to prevent overfitting and increases robustness.

    \item \emph{Sequential Feature Selection:}  
    Sequential FS is a wrapper method that repetitively adds or removes features based on their effect on classifier performance\cite{kohavi1997wrappers}. It selects features by adding one feature at a time (forward selection) or removing one feature at a time (backward elimination). This terminates when a predefined number of features is reached or when performance no longer improves. It is computationally intensive but can capture feature interactions and optimize the feature subset for specific classifiers.

\end{itemize}
\subsubsection{Feature Selection using LLM-FS}
We propose LLM-FS, a zero-shot feature selection framework that leverages LLMs to score the importance of each feature based on its statistical properties, class-conditional distributions, and representative values along with task context. The traditional FS methods primarily rely on statistical heuristics or model-driven scores, often overlooking the semantic context of features. While LLM-FS integrates descriptive statistics with the semantic reasoning abilities of pretrained LLMs.  LLM-FS operates by constructing structured prompts for each feature, which include its statistical descriptors and the classification task context. The LLM then processes these prompts to generate an importance score for each feature, reflecting its relevance to the malware detection task. This approach allows the LLM to leverage its vast pretraining knowledge to infer feature importance, even in the absence of raw data distributions. The LLM-FS feature scoring algorithm is summarized in Algorithm \ref{alg:llm_scoring_clean}.\\
Formally, for each feature $f_j$ ($j=1,\dots,d$), we compute the following statistics:
\begin{itemize}
    \item \textbf{Global statistics:} mean($\mu_j$), variance($\sigma_j$), median, min, max, interquartile range (IQR).  
    \item \textbf{Class-conditional statistics:} class-wise means $\mu_j^{(1)}, \mu_j^{(0)}$, standard deviations $\sigma_j^{(1)}, \sigma_j^{(0)}$, and mean difference ($\Delta \mu_j$).  
\end{itemize}
The computation of these statistics are defined as follows:
\begin{equation}
\mu_j = \frac{1}{n} \sum_{i=1}^{n} X_{ij}
\end{equation}
\begin{equation}
\sigma_j = \sqrt{\frac{1}{n} \sum_{i=1}^{n} (X_{ij} - \mu_j)^2}
\end{equation}
\begin{equation}
\mu_j^{(1)} = \frac{1}{n_1} \sum_{i:y_i=1} X_{ij}
\end{equation}
\begin{equation}
 \mu_j^{(0)} = \frac{1}{n_0} \sum_{i:y_i=0} X_{ij}
\end{equation}
\begin{equation}
\sigma_j^{(1)} = \sqrt{\frac{1}{n_1} \sum_{i:y_i=1} (X_{ij} - \mu_j^{(1)})^2}
\end{equation}
\begin{equation}
\sigma_j^{(0)} = \sqrt{\frac{1}{n_0} \sum_{i:y_i=0} (X_{ij} - \mu_j^{(0)})^2} 
\end{equation}
\begin{equation}
\Delta \mu_j = \mu_j^{(1)} - \mu_j^{(0)}
\end{equation}
where \(n_1\) and \(n_0\) denote the number of samples in the malware and benign classes, respectively. Using these statistics, a feature descriptor $\mathcal{D}(f_j)$ is constructed, which is then embedded into a structured prompt $P_j$ along with the task context($\mathcal{C}$).  $\mathcal{C}$ specifies the classification objective (e.g., "classify whether a given file is malware or benign").
The feature descriptor for $f_j$ is defined as:
\begin{equation}
\begin{aligned}
\mathcal{D}(f_j) = \{&\mu_j, \sigma_j, \text{median}, \min, \max, \text{IQR},\\
&\mu_j^{(1)}, \mu_j^{(0)}, \sigma_j^{(1)}, \sigma_j^{(0)}, 
\Delta \mu_j, \text{samples}\}.
\end{aligned}
\end{equation}
Then the LLM maps this descriptor and task context into an importance score $s_j$.
\begin{equation}
  s_j = \Phi(\mathcal{D}(f_j), \mathcal{C}), \quad s_j \in [0,1]
\end{equation}
The $s_j$ values are aggregated to form a score vector $S = (s_1, s_2, \dots, s_d)$, which is used to rank features for selection. If the value of $s_j$ is closer to 1, it indicates that the feature $f_j$ is highly relevant for distinguishing between malware and benign samples. Conversely, if $s_j$ is closer to 0, it suggests that the feature is irrelevant. Intermediate values (e.g., around 0.5) indicate moderate relevance. The top-$k$ features are then selected based on these scores for training classifiers. In this way, LLM-FS combines quantitative separation derived from descriptive statistics with qualitative reasoning encoded in LLM knowledge. This dual perspective postions LLM-FS as a competitive alternative to traditional FS methods, while introducing semantic interpretability.

        


\begin{algorithm}
\caption{LLM-FS Feature Scoring Algorithm}
\label{alg:llm_scoring_clean}
\KwIn{$X \in \mathbb{R}^{n \times d}$ (feature matrix), $y \in \{0,1\}^n$ (labels)}
\KwOut{$S = (s_1,\dots,s_d)$, where $s_j\in[0,1]$ (feature importance scores)}

\For{$j \gets 1$ \textbf{to} $d$}{
  \begin{enumerate}
    \item \textbf{Compute statistics} for $f_j$: 
    \begin{itemize}
      \item Global — mean $(\mu_j)$, standard deviation $(\sigma_j)$, median, min, max, interquartile range (IQR).
      \item Class-conditional — class-wise means $(\mu_j^{(1)}, \mu_j^{(0)})$, standard deviations $(\sigma_j^{(1)}, \sigma_j^{(0)})$,\\
      and mean difference $\Delta\mu_j = \mu_j^{(1)} - \mu_j^{(0)}$.
    \end{itemize}

    \item \textbf{Construct feature descriptor} $\mathcal{D}(f_j)$ \\

    \item \textbf{Formulate structured prompt} $P_j = \mathcal{C} \parallel \mathcal{D}(f_j)$, 
    where $\mathcal{C}$ denotes the task context 

    \item \textbf{Query LLM} with $P_j$ using deterministic decoding (temperature $=0$) to obtain raw output $o_j$.

    \item \textbf{Parse and validate response:}\\
    \eIf{$o_j$ is a valid numeric value and $o_j \in [0,1]$}
    {
      $s_j \leftarrow o_j$ \tcp*{accept score}
    }
    {
      $s_j \leftarrow 0.5$ \tcp*{fallback neutral score}
    }
  \end{enumerate}
}
\Return $S = (s_1,\dots,s_d)$
\end{algorithm}


\subsection{Classification}
LLM-FS employs multiple classifiers to evaluate the effectiveness of the selected feature subsets. We utilize four widely used classifiers: Random Forest (RF), Extra Trees (ET), Multilayer Perceptron (MLP), and k-Nearest Neighbors (KNN)\cite{singh2021survey,aslan2020comprehensive}. These classifiers were chosen for their diverse learning paradigms and proven effectiveness in malware detection tasks. Each classifier is trained on the feature subsets selected by both traditional FS methods and LLM-FS, allowing for a comprehensive comparison of classification performance across different feature selection strategies.
\subsubsection{Random Forest (RF)}
Random Forest (RF) is a robust ensemble learning technique that combines multiple decision trees to enhance classification performance. Each tree in the forest is trained on a bootstrap sample of the data, and at each split, a random subset of features is considered, promoting diversity among the trees. It aggregates the predictions of individual trees by majority voting for classification tasks.

\subsubsection{Extra Trees (ET)}
Extra Trees (ET) is an ensemble learner that also builds upon decision trees. It introduces additional randomness by selecting split points randomly for each feature, further reducing variance and enhancing model robustness. ET are computationally efficient and often achieve comparable or superior performance to Random Forest.
\subsubsection{Multilayer Perceptron (MLP)}
Multilayer Perceptron is a class of feedforward artificial neural networks that consist of multiple layers of interconnected neurons. MLPs are capable of learning complex, non-linear decision boundaries through backpropagation and gradient descent optimization. They are particularly effective in scenarios where the relationship between features and the target variable is intricate and not easily captured by linear models. MLPs can automatically learn feature representations, making them suitable for high-dimensional data. However, they require careful tuning of hyperparameters such as the no. of layers, neurons per layer, learning rate, and regularization techniques to prevent overfitting.
\subsubsection{k-Nearest Neighbors (KNN)}
k-Nearest Neighbors is a simple yet effective, non-parametric and instance-based learning algorithm that classifies samples based on the majority class of their k-nearest neighbors in the feature space. It works based on the principle that similar instances tend to belong to the same class. The neighborhood is typically measured using distance metrics such as Euclidean or Manhattan distance. It often performs well in scenarios where the decision boundary is irregular and can adapt to local data structures.
\section{Experiments and Evaluation}
\subsection{Experimental Setup}
The experiments were done using NVIDIA DGX A100, a high-performance computing system specifically built for AI workloads. The system has a 64-core processor (AMD EPYC 7742), 512GB of DDR4 RAM, with 2 NVIDIA A100 GPUs (each with 40GB of memory). This sturdy hardware configuration make sure the efficient handling of computational demands which are associated with training many classifiers and performing feature selection on the EMBOD dataset. The LLM-FS framework was implemented using Python, and various LLMs were brought to use via their respective APIs for feature scoring. The OpenAI API was used for accessing GPT-5.0-mini, GPT-4.0, and GPT-4.0-mini models, while the Google gen-ai API provided access to Gemini-2.5-flash and Gemini-2.5-pro models. 
Default hyperparameters were involved for all classifiers to make sure a fair unbiased comparison across various FS methods. 


\subsection{Dataset}
\begin{figure}[h!]
\centering
  \includegraphics[scale=0.40]{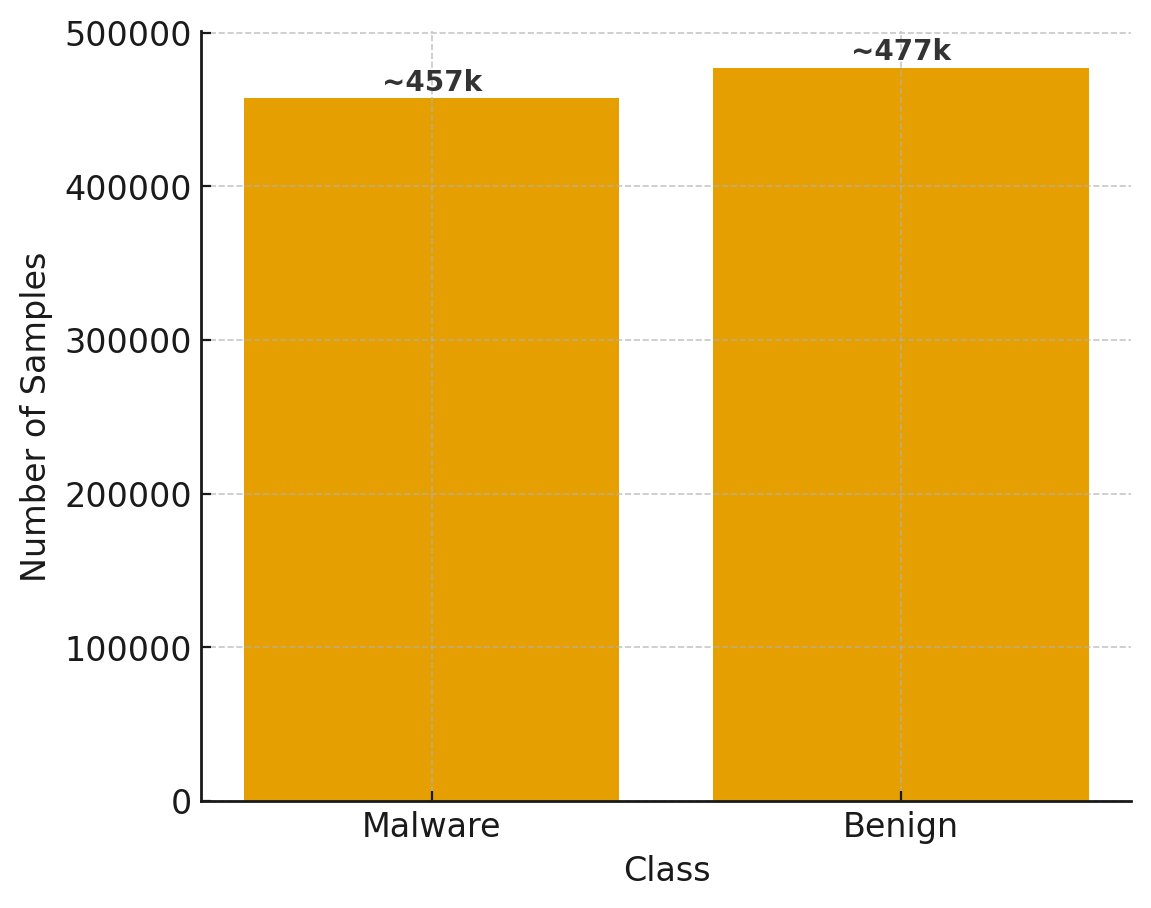}
  \caption{Distribution of samples in the EMBOD dataset}
  \label{fig:embod-distribution}
\end{figure}

The EMBOD\footnote{https://www.kaggle.com/datasets/ajvadhaneef/embod-all} malware dataset has been used for judging the performance of the proposed LLM-FS paradigm. EMBOD dataset is a mix of EMBER\cite{anderson2018ember} and BODMAS\cite{yang2021bodmas} benchmark datasets, giving a thorough collection of malware samples for machine learning and deep learning tasks. It has a total \~934k samples,  offering a rich foundation for training and evaluating different classifiers. Every sample in the dataset is depicted as a 2381-dimensional feature vector extracted using the LIEF project\footnote{https://github.com/lief-project/LIEF} (version 0.9.0). This feature vector has features such as headers, imports, sections, histograms, entropy statistics, and strings.  The spread of samples in the dataset is depicted in Figure \ref{fig:embod-distribution}. The dataset was split into training and testset with 80:20 ratio.

\subsection{Evaluation Metrics\label{sec:metrics}}
LLM-FS was tested using commonly used metrics, such as accuracy, precision, recall, and F1-Score. Other relevant metrics, such as AUC and MCC, were also considered to calculate the effectiveness of LLM-FS, as they provide better understanding of the model's performance in uncertain situations. The main metrics are represented in terms of TP, TN, FP, and FN for every class. The $T_c$ shows the time taken for the feature selection, model training, and evaluation, meant to give insights into the computational efficiency of all FS methods. The Table \ref{tab:metrics} summarizes the formulae for each metric.


  \begin{table}[!ht]
  \small
    \centering
    \begin{threeparttable}
    \caption{Evaluation Metrics}
    \label{tab:metrics}
    \begin{tabular}{ll}
        \toprule
        \textbf{Metric} & \textbf{Formula} \\
          \midrule
        Accuracy  ($A_c$) & $\frac{TP + TN}{TP + TN + FP + FN} $ \\
        \midrule
        Precision ($P_c$) & $\frac{TP}{TP+FP} $ \\
        \midrule 
        Recall   ($R_c$) & $\frac{TP}{TP+FN} $ \\
        \midrule
        F1-Score ($F_1$) & $2 \cdot \frac{P_c \cdot R_c}{P_c + R_c} $ \\ \midrule
        AUC ($U_c$) & $\int_{0}^{1} TPR \, d(FPR) $ \\ \midrule
        MCC ($M_c$) & $\frac{(TP \cdot TN) - (FP \cdot FN)}{\sqrt{(TP + FP)(TP + FN)(TN + FP)(TN + FN)}} $ \\
        \midrule
        Total Runtime ($T_c$) & $T_2 - T_1$ \\
        \bottomrule
    \end{tabular}
    \begin{tablenotes}
        \footnotesize
        \item $TP$: True Positives, $TN$: True Negatives, $FP$: False Positives, $FN$: False Negatives, $TPR$: True Positive Rate, $FPR$: False Positive Rate, $T_1$: Start Time, $T_2$: End Time
      \end{tablenotes}
    \end{threeparttable}
  \end{table}

\subsection{Comparison of LLM-based and Traditional FS Methods}
\begin{center}
\begin{figure*}[!ht]
\centering
\includegraphics[width=.75\textwidth]{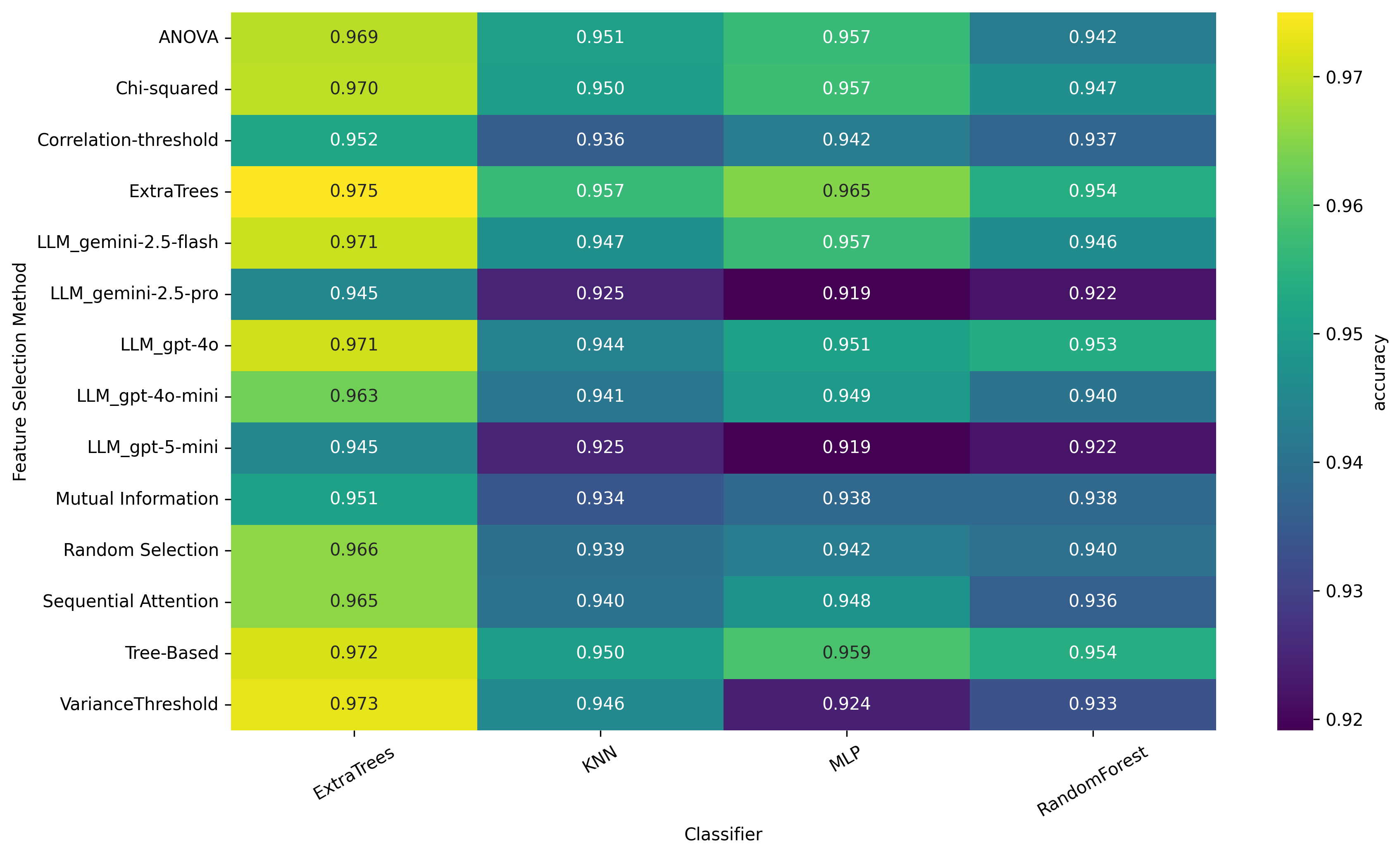}
\caption{Heatmap comparing LLM-based and traditional FS Methods across multiple classifiers: (a) Random Forest, (b) Extra Trees (c) KNN , and (d) MLP. Each cell represents the accuracy achieved by a specific combination of feature selection method and classifier, with color intensity indicating performance level.}
\label{fig:heatmap}
\end{figure*}
\end{center}

The performance comparison of LLM-based and traditional feature selection methods across multiple classifiers is summarized in Table \ref{tab:comparison}. The table presents key metrics including the number of selected features ($N_c$), $A_c$, $P_c$, $R_c$, $F_1$, $U_c$, $M_c$, and total runtime in seconds ($T_c$) for each combination of feature selection method and classifier. We also provide a heatmap visualization in Figure \ref{fig:heatmap} to illustrate the performance differences across methods and classifiers. Traditional methods such as Variance Threshold, Tree-Based importance, Sequential Attention, Random Selection, Mutual Information, Extra Trees, Correlation-threshold, Chi-Squared, and ANOVA are compared against LLM-based methods using GPT-5.0-mini, GPT-4.0-mini, GPT-4.0, Gemini-2.5-pro, and Gemini-2.5-flash. Top-341 features were selected based on the importance scores from each FS method, and the classifiers were trained and evaluated using these subsets. The 341 features were chosen based on preliminary experiments that indicated this number provided a good balance between dimensionality reduction and model performance. 

\begin{table*}[!htbp]
\small
  \centering
  \begin{threeparttable}
  \caption{Performance Comparison of LLM-based and Traditional Feature Selections with Multiple Classifiers}
  \label{tab:comparison}
  \begin{tabular}{llcccccccc}
      \toprule
     \textbf{FS Method} & \textbf{Classifier} & \textbf{$N_c$} & \textbf{$A_c$} & \textbf{$P_c$} & \textbf{$R_c$} & \textbf{$F_1$} & \textbf{$U_c$} & \textbf{$M_c$} & \textbf{$T_c$} \\ \midrule
     VarianceThreshold & RandomForest & 341 & 0.933 & 0.935 & 0.933 & 0.933 & 0.987 & 0.868 & 2091.934 \\
     Tree-Based & RandomForest & 341 & 0.954 & 0.954 & 0.954 & 0.954 & 0.992 & 0.908 & 1309.516 \\
     Sequential Attention & RandomForest & 341 & 0.936 & 0.936 & 0.936 & 0.936 & 0.985 & 0.872 & 260.015 \\
     Random Selection & RandomForest & 341 & 0.94 & 0.94 & 0.94 & 0.94 & 0.986 & 0.879 & 269.626 \\
     Mutual Information & RandomForest & 341 & 0.938 & 0.938 & 0.938 & 0.938 & 0.986 & 0.876 & 1567.062 \\
     ExtraTrees & RandomForest & 341 & 0.954 & 0.954 & 0.954 & 0.954 & 0.992 & 0.907 & 628.826 \\
     Correlation-threshold & RandomForest & 341 & 0.937 & 0.937 & 0.937 & 0.937 & 0.986 & 0.874 & 863.982 \\
     Chi-squared & RandomForest & 341 & 0.947 & 0.947 & 0.947 & 0.947 & 0.99 & 0.893 & 734.576 \\
     ANOVA & RandomForest & 341 & 0.942 & 0.942 & 0.942 & 0.942 & 0.988 & 0.885 & 722.611 \\
     LLM\_gpt-5-mini & RandomForest & 341 & 0.922 & 0.922 & 0.922 & 0.922 & 0.978 & 0.844 & 765.264 \\
     LLM\_gpt-4o-mini & RandomForest & 341 & 0.94 & 0.94 & 0.94 & 0.94 & 0.987 & 0.881 & 725.55 \\
     LLM\_gpt-4o & RandomForest & 341 & 0.953 & 0.954 & 0.953 & 0.953 & 0.991 & 0.907 & 9698.588 \\
     LLM\_gemini-2.5-pro & RandomForest & 341 & 0.922 & 0.922 & 0.922 & 0.922 & 0.978 & 0.844 & 798.172 \\
     LLM\_gemini-2.5-flash & RandomForest & 341 & 0.946 & 0.946 & 0.946 & 0.946 & 0.99 & 0.892 & 650.018 \\
     \midrule
     VarianceThreshold & MLP & 341 & 0.924 & 0.924 & 0.924 & 0.924 & 0.976 & 0.848 & 2199.648 \\
     Tree-Based & MLP & 341 & 0.959 & 0.959 & 0.959 & 0.959 & 0.989 & 0.918 & 3235.924 \\
     Sequential Attention & MLP & 341 & 0.948 & 0.948 & 0.948 & 0.948 & 0.984 & 0.895 & 4233.249 \\
     Random Selection & MLP & 341 & 0.942 & 0.942 & 0.942 & 0.942 & 0.983 & 0.884 & 3474.474 \\
     Mutual Information & MLP & 341 & 0.938 & 0.938 & 0.938 & 0.938 & 0.979 & 0.876 & 5444.09 \\
     ExtraTrees & MLP & 341 & 0.965 & 0.965 & 0.965 & 0.965 & 0.993 & 0.929 & 3595.751 \\
     Correlation-threshold & MLP & 341 & 0.942 & 0.942 & 0.942 & 0.942 & 0.978 & 0.884 & 8090.274 \\
     Chi-squared & MLP & 341 & 0.957 & 0.957 & 0.957 & 0.957 & 0.989 & 0.915 & 4872.486 \\
     ANOVA & MLP & 341 & 0.957 & 0.957 & 0.957 & 0.957 & 0.989 & 0.914 & 4591.393 \\
     LLM\_gpt-5-mini & MLP & 341 & 0.919 & 0.919 & 0.919 & 0.919 & 0.97 & 0.838 & 2935.147 \\
     LLM\_gpt-4o-mini & MLP & 341 & 0.949 & 0.949 & 0.949 & 0.949 & 0.984 & 0.898 & 4328.015 \\
     LLM\_gpt-4o & MLP & 341 & 0.951 & 0.951 & 0.951 & 0.951 & 0.986 & 0.902 & 5231.601 \\
     LLM\_gemini-2.5-pro & MLP & 341 & 0.919 & 0.919 & 0.919 & 0.919 & 0.97 & 0.838 & 2341.13 \\
     LLM\_gemini-2.5-flash & MLP & 341 & 0.957 & 0.957 & 0.957 & 0.957 & 0.988 & 0.914 & 4977.999 \\
     \midrule
     VarianceThreshold & KNN & 341 & 0.946 & 0.946 & 0.946 & 0.946 & 0.98 & 0.891 & 20343.59 \\
     Tree-Based & KNN & 341 & 0.95 & 0.95 & 0.95 & 0.95 & 0.983 & 0.9 & 118.482 \\
     Sequential Attention & KNN & 341 & 0.94 & 0.94 & 0.94 & 0.94 & 0.978 & 0.88 & 111.201 \\
     Random Selection & KNN & 341 & 0.939 & 0.939 & 0.939 & 0.939 & 0.979 & 0.879 & 111.148 \\
     Mutual Information & KNN & 341 & 0.934 & 0.934 & 0.934 & 0.934 & 0.975 & 0.868 & 120.869 \\
     ExtraTrees & KNN & 341 & 0.957 & 0.957 & 0.957 & 0.957 & 0.986 & 0.914 & 59.677 \\
     Correlation-threshold & KNN & 341 & 0.936 & 0.936 & 0.936 & 0.936 & 0.976 & 0.871 & 113.604 \\
     Chi-squared & KNN & 341 & 0.95 & 0.95 & 0.95 & 0.95 & 0.982 & 0.9 & 113.412 \\
     ANOVA & KNN & 341 & 0.951 & 0.951 & 0.951 & 0.951 & 0.983 & 0.901 & 110.396 \\
     LLM\_gpt-5-mini & KNN & 341 & 0.925 & 0.925 & 0.925 & 0.925 & 0.969 & 0.849 & 64.756 \\
     LLM\_gpt-4o-mini & KNN & 341 & 0.941 & 0.941 & 0.941 & 0.941 & 0.978 & 0.882 & 61.496 \\
     LLM\_gpt-4o & KNN & 341 & 0.944 & 0.944 & 0.944 & 0.944 & 0.98 & 0.888 & 21135.485 \\
     LLM\_gemini-2.5-pro & KNN & 341 & 0.925 & 0.925 & 0.925 & 0.925 & 0.969 & 0.849 & 58.516 \\
     LLM\_gemini-2.5-flash & KNN & 341 & 0.947 & 0.947 & 0.947 & 0.947 & 0.982 & 0.894 & 60.413 \\
     \midrule
     VarianceThreshold & ExtraTrees & 341 & 0.973 & 0.973 & 0.973 & 0.973 & 0.996 & 0.946 & 9026.76 \\
     Tree-Based & ExtraTrees & 341 & 0.972 & 0.972 & 0.972 & 0.972 & 0.996 & 0.943 & 377.077 \\
     Sequential Attention & ExtraTrees & 341 & 0.965 & 0.966 & 0.965 & 0.965 & 0.995 & 0.931 & 231.015 \\
     Random Selection & ExtraTrees & 341 & 0.966 & 0.966 & 0.966 & 0.965 & 0.995 & 0.931 & 229.014 \\
     Mutual Information & ExtraTrees & 341 & 0.951 & 0.951 & 0.951 & 0.951 & 0.991 & 0.902 & 618.389 \\
     ExtraTrees & ExtraTrees & 341 & 0.975 & 0.975 & 0.975 & 0.975 & 0.997 & 0.95 & 200.604 \\
     Correlation-threshold & ExtraTrees & 341 & 0.952 & 0.953 & 0.952 & 0.952 & 0.991 & 0.905 & 211.899 \\
     Chi-squared & ExtraTrees & 341 & 0.97 & 0.97 & 0.97 & 0.97 & 0.996 & 0.939 & 174.918 \\
     ANOVA & ExtraTrees & 341 & 0.969 & 0.969 & 0.969 & 0.969 & 0.996 & 0.938 & 176.325 \\
     LLM\_gpt-5-mini & ExtraTrees & 341 & 0.945 & 0.945 & 0.945 & 0.945 & 0.988 & 0.89 & 229.463 \\
     LLM\_gpt-4o-mini & ExtraTrees & 341 & 0.963 & 0.963 & 0.963 & 0.963 & 0.994 & 0.926 & 211.524 \\
     LLM\_gpt-4o & ExtraTrees & 341 & 0.971 & 0.971 & 0.971 & 0.971 & 0.996 & 0.942 & 6892.361 \\
     LLM\_gemini-2.5-pro & ExtraTrees & 341 & 0.945 & 0.945 & 0.945 & 0.945 & 0.988 & 0.89 & 238.156 \\
     LLM\_gemini-2.5-flash & ExtraTrees & 341 & 0.97 & 0.971 & 0.97 & 0.97 & 0.996 & 0.941 & 191.902 \\
    
      \bottomrule
  \end{tabular}
  \begin{tablenotes}
      \footnotesize
      \item $N_c$: Number of features selected; $A_c$: Accuracy; $P_c$: Precision; $R_c$: Recall; $F_1$: F1-Score; $U_c$: AUC; $M_c$: MCC; $T_c$: Total time taken (in seconds) for feature selection, model training and evaluation.
    \end{tablenotes}
  \end{threeparttable}
\end{table*}

\subsection{Discussion}
 Results indicate that zero-shot LLM-guided feature selection (LLM-FS) is a competitive alternative to classical filter, wrapper, and embedded approaches on the EMBOD benchmark. Across four classifiers (RF, ET, MLP, KNN) and multiple evaluation metrics ($A_c$, F1, AUC, MCC), LLM-FS consistently matched or outperformed strong baselines while requiring fewer assumptions about the underlying data distribution. Importantly, the features prioritized by LLM-FS frequently aligned with domain-relevant priors in malware analysis—such as entropy- and section-related descriptors, import usage, and header-level indicators—supporting the claim that LLMs can inject semantic knowledge into tabular FS. A second key observation is the \emph{classifier-agnostic behavior} of LLM-FS. The relative ranking of FS methods was broadly stable across ensemble-based and non-ensemble learners, suggesting that the advantages of LLM-FS are not tied to a particular model family. This property makes it practical in operational settings, as the same feature subset can be reused across different classifiers without the need for repeated feature selection.  Third, LLM-FS contributes to \emph{interpretability}. Beyond producing numerical scores, LLMs are capable of providing natural-language rationales that link statistical separability to meaningful behavioral hypotheses (e.g., ``elevated section entropy may indicate packing/obfuscation''). Even without explicit rationales, the strong alignment between top-ranked features and established malware indicators increases analyst trust compared to purely statistical scores.\\
We further examine the \emph{stability and robustness}. While wrapper-based methods can fluctuate under resampling, LLM-FS showed competitive or lower variance across runs and across different LLM backbones, particularly when deterministic decoding (temperature = 0) and fixed-order prompt templates were used. However, some variability persists due to model updates and API nondeterminism. Ensembling across prompts or averaging repeated scores helps mitigate this variance. Additionally, we analyzed the fallback rule, where invalid or non-numeric outputs are replaced with a neutral score of $s = 0.5$. This mechanism improves robustness by preventing failure cases but may slightly dilute rankings when many features trigger fallbacks; in practice, its impact was minimal given the high rate of valid outputs. Regarding \emph{efficiency}, LLM-FS introduces computational overhead during the feature scoring stage due to API latency and per-feature queries. Once the feature subset is determined, however, downstream training and inference remain as efficient as with any reduced feature set. This cost can be amortized by performing FS periodically (e.g., per dataset refresh) rather than per training run. In constrained environments, a hybrid strategy—combining fast statistical prefilters (Variance Threshold, Mutual Information) with LLM-based re-ranking—achieves much of the interpretability benefit while reducing API costs. Finally, we highlight two \emph{threats to validity}. First, when using class-conditional statistics in prompts, these must be computed strictly on the training split to avoid label leakage. Second, fusing EMBER and BODMAS introduces potential distributional shifts; cross-dataset and temporal evaluations would further strengthen conclusions. Moreover, the effectiveness of LLM-FS depends on the semantic quality of feature names; obfuscated or synthetic features may reduce its zero-shot advantage.\\
Although LLM-FS gets strong competitive results, several aspects need deeper investigation. The semantic richness of feature names or descriptors widely affect the framework’s performance, which may get negatively affected when features are purely numeric or vague. Future versions could make use of embedding-based representations or automatically generated textual substitutes to get over this upper bound. Scalability is one such concern, as running one prompt per feature becomes computationally expensive for very high-dimensional data; hybrid pipelines that inculcate fast statistical filters before LLM re-ranking can give an efficient and cheap solution. Moreover, adversarial robustness is yet to be researched upon, particularly how deliberately fabricated or misleading feature names might affect LLM reasoning, a key component for real-world security applications. The comparative scope of this work currently stresses on traditional feature selection baselines; inculcating deep and reinforcement learning-based techniques would offer a better evaluation. Finally, even though the LLM-FS scoring stage brings some latency, it is performed offline and reused during retraining. Future research may explore lighter, distilled variants to improve runtime efficiency and real-life implementation.

\section{Conclusion}
LLM-FS gives a zero-shot, knowledge-based way for feature selection in malware detection,  designed particularly for extensive and sophisticated tabular datasets. The framework makes use of LLMs to better understand structured feature descriptors that combine global and class-level statistics. This lowers reliability on statistical filters or heuristics that are associated with a specific model. This enables the system to give more meaningful scores based on logic rather than just looking at raw data. Experiments on the EMBOD malware dataset, which involved several classifiers and performance metrics, show that LLM-FS is as accurate as or more accurate than traditional methods like Variance Threshold, $\chi^2$, ANOVA, Mutual Information, Tree-based models, ExtraTrees, and Sequential Selection. Not just a good predictive performance, the framework also gives consistent and interpretable feature subsets, decreased reliance on labeled data, and stable behavior across runs. Although the technique introduces some extra computational expenses and depends on vivid feature naming, these challenges can be dealt with through hybrid strategies that mix classical filtering with LLM-based refinement. Overall, LLM-FS demonstrates that LLMs can effectively reduce the gap between statistical feature selection and semantic reasoning, making malware-detection systems more transparent, label-efficient, and reliable.

\bibliographystyle{ieeetr}
\bibliography{refs}







\end{document}